\documentclass[letterpaper, 10 pt, conference]{ieeeconf}  

\IEEEoverridecommandlockouts                              

\usepackage{algorithmicx}
\usepackage{algpseudocode}
\usepackage{graphicx}
\usepackage{hyperref}
\usepackage{xcolor}
\usepackage{bbding}
\usepackage{colortbl}
\usepackage{tabularx}
\usepackage{threeparttable}
\usepackage[linesnumbered, ruled]{algorithm2e}
\usepackage{amsmath}
\usepackage{booktabs}

\title{\LARGE \bf
Navigation with VLM framework:\\ Towards Going to Any Language
}

\author{Zecheng Yin$^{1,4}$, Chonghao Cheng$^{2}$, Yao Guo$^{3}$, Zhen Li$^{1,4}$
\thanks{$^{1}$ Future Network of Intelligence Institute(Shenzhen),
        Shenzhen, China
        {\tt\small yinzecheng.cuhk@gmail.com}}%
\thanks{$^{2}$ University of Technology Sydney, Sydney, Australia
        {\tt\small chonghaocheng@student.uts.edu.au}}%
\thanks{$^{3}$Shanghai Jiao Tong University, Shanghai, China
        {\tt\small }}%
\thanks{$^{4}$The Chinese University of Hongkong(Shenzhen), Shenzhen, China
        {\tt\small lizhen@cuhk.edu.cn}}%
}

\begin{document}

\maketitle
\thispagestyle{empty}
\pagestyle{empty}

\begin{abstract}
Navigating towards fully open language goals and exploring open scenes in an intelligent way have always raised significant challenges. Recently, Vision Language Models (VLMs) have demonstrated remarkable capabilities to reason with both language and visual data. Although many works have focused on leveraging VLMs for navigation in open scenes, they often require high computational cost, rely on object-centric approaches, or depend on environmental priors in detailed human instructions.
We introduce Navigation with VLM (NavVLM), a training-free framework that harnesses open-source VLMs to enable robots to navigate effectively, even for human-friendly language goal such as abstract places, actions, or specific objects in open scenes. NavVLM leverages the VLM as its cognitive core to perceive environmental information and constantly provides exploration guidance achieving intelligent navigation with only a neat target rather than a detailed instruction with environment prior. 
We evaluated and validated NavVLM in both simulation and real-world experiments. In simulation, our framework achieves state-of-the-art performance in Success weighted by Path Length (SPL) on object-specifc tasks in richly detailed environments from Matterport 3D (MP3D), Habitat Matterport 3D (HM3D) and Gibson. With navigation episode reported, NavVLM demonstrates the capabilities to navigate towards any open-set languages. In real-world validation, we validated our framework's effectiveness in real-world robot at indoor scene. Our code and data are publicly available.

\end{abstract}


\begin{figure}[t]
\centering

\begin{tabular}{c}
  \includegraphics[width=0.9\linewidth]{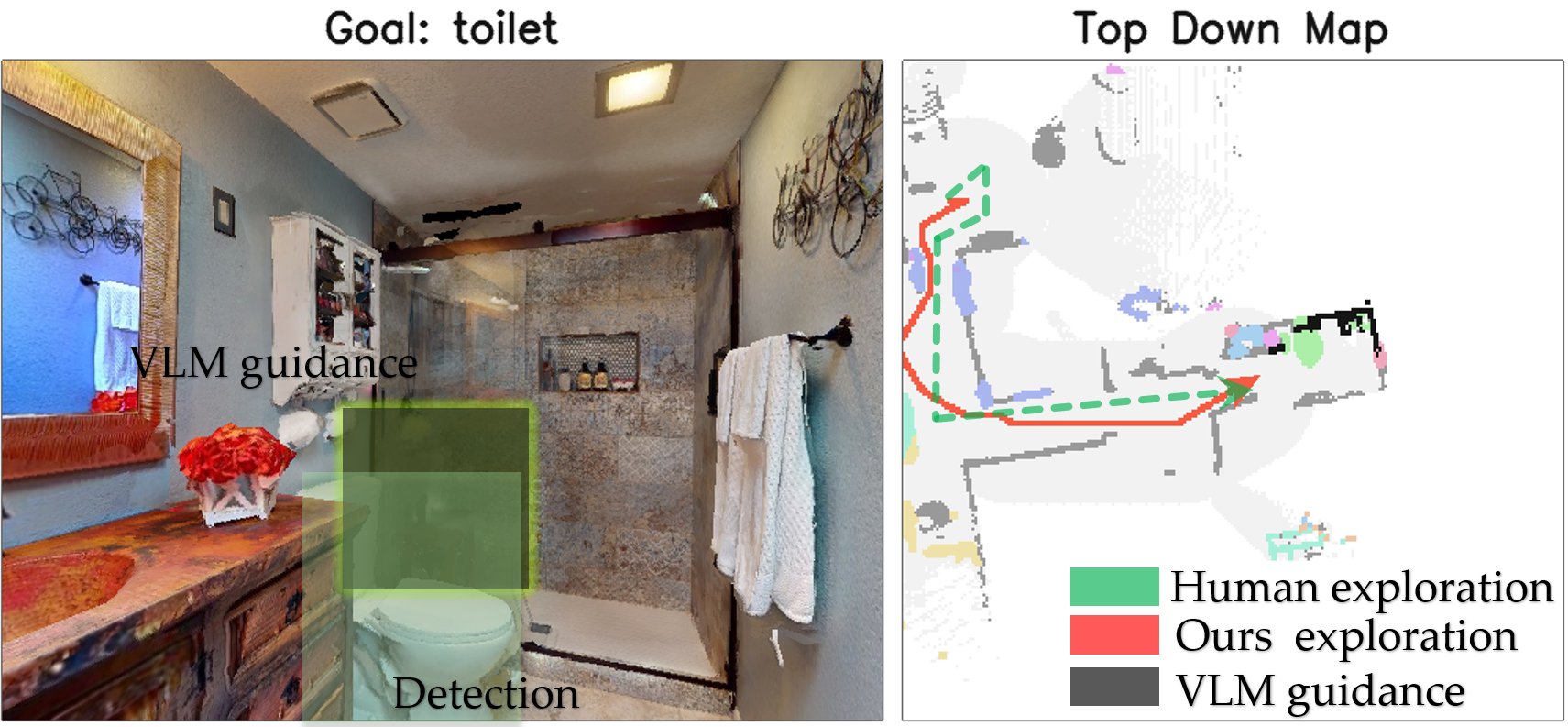} \\
  (a) \\[1em]
  \includegraphics[width=0.9\linewidth]{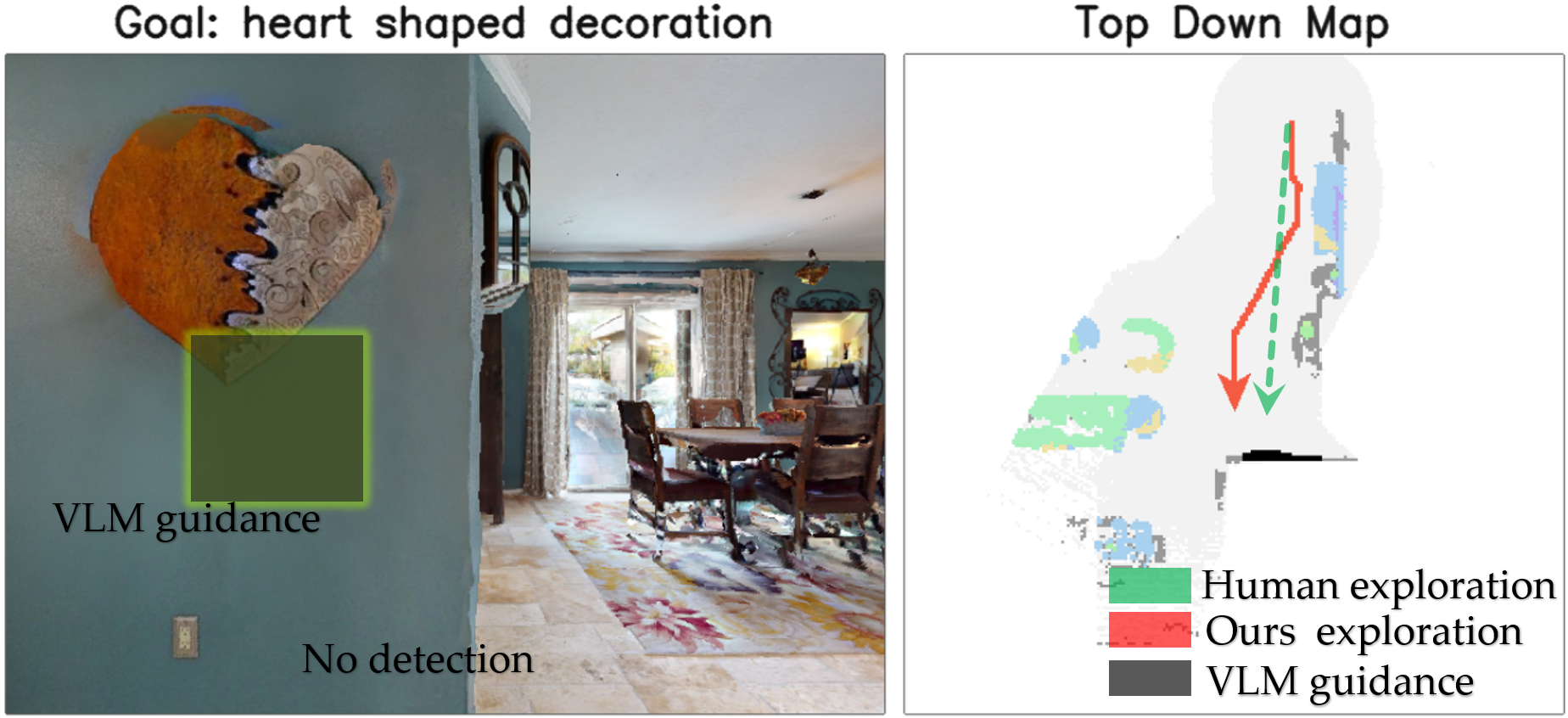} \\
  (b) \\[1em]
  \includegraphics[width=0.9\linewidth]{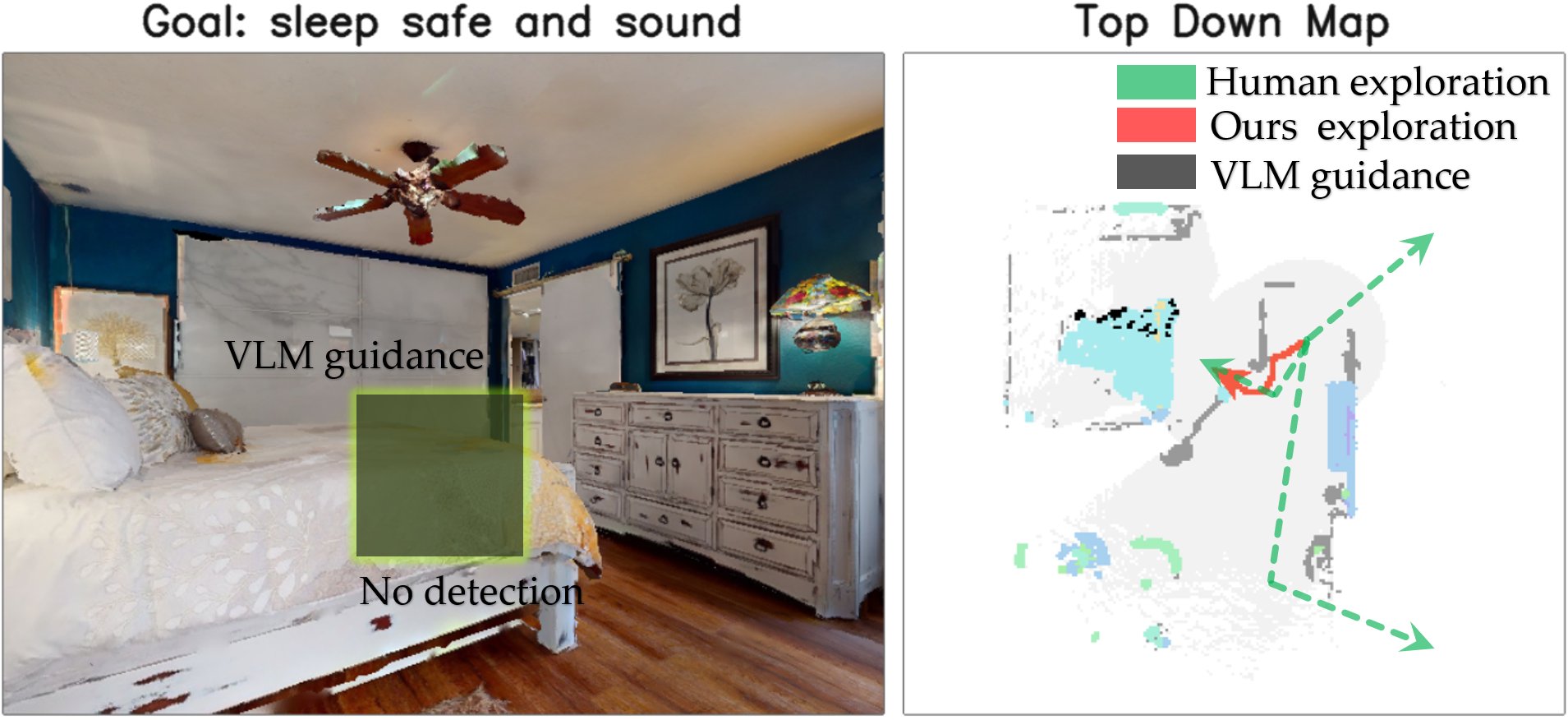} \\
  (c)
\end{tabular}

\caption{
  (a) NavVLM can perform intelligent exploration in open scenes for object-centric goals.
  (b) NavVLM can guide to out-of-domain language goals.
  (c) NavVLM can navigate to any human-friendly open goals such as action, or even abstract places.
}
\label{fig:all_images}
\end{figure}
\section{Introduction}
Object-oriented Naviagtion (ObjNav) is to autonomously find specific objects in unknown environments\cite{objnav}, where the robot actively explores the unknown environment with the purpose of finding the target and navigates to it. This task puts challenges on the intelligence of the active exploration, effectiveness of navigation, and  vision-language understanding. This problem combines exploration and navigation in one phase, rather than navigation after exploration. \textbf{ObjNav is not task} like Vison Language Navigation in Continuous Environment (VLN-CE)\cite{vlnce}, where the human language instructions contain environmental information. ObjNav task requires the robot to explore the environment and capture the clues itself to find the target.

The vision understanding of RGB observation is a key factor for navigation performance\cite{vln_acl}. Vision Language Models (VLMs)\cite{li2024llavanext-strong,internvl} are more focused on text-driven tasks that interact with RGB\cite{llm_eval,llmseg}, such as vison-question answering (VQA)\cite{lisa++}, knowledge completion, which is closely related to vision-based object-oriented navigation, demonstrating potential capabilities of serving the cognitive core of the agent and guide navigation with consensus\cite{llm_survey}.

Some more works extend navigation goal language with more open vocabulary abilities\cite{conceptgraph} and some leverage VLMs \cite{kuang2024openfmnav} for object-oriented navigation, but they either limit their methods to object-centric items or languages implicitly targeted on object such as "table", "places for sleeping",  or require human conversation intervention for satisfying performance\cite{findthis,shah2023lfg}. Language goals not anchored onto objects like "center in the hall", "a corner in kitchen" are not applicable for these object-centric works. In the real world, the need for mobility assistance is more likely related to ambiguous languages such as a daily action, location, or even abstract place, which are out of the domain of current object-centric items.


To move beyond the object-anchored limitation and make autonomous navigation more reflective of real-world scenarios, extended on ObjNav, we propose \textbf{Open Goal Navigation Problem}.\label{full_open}
For an agent initialized in random location in an open scene without prior knowledge, the open set navigation is to let the agent navigate to a language goal directly and autonomously. The navigation purpose of open goal navigation can be wilder, more ambiguous and abstract, such as ``a corner in the kitchen", ``somewhere I can eat dinner", or an abstract place ``the middle of the hall", which are more friendly to daily conversation and clearly out-of-domain of object-centric systems.

\begin{table}[h]
\caption{Open Navigation Language Goal.}\label{tab1}
\centering

\begin{tabular}{|c|c|c|c|}
\hline
Language Goal &  Expl.+Detect. & Recent & Ours\\
\hline
apple &  \CheckmarkBold & \CheckmarkBold& \CheckmarkBold \\
apple on desk &  \XSolidBold & \CheckmarkBold & \CheckmarkBold\\
cook dinner & \XSolidBold &\CheckmarkBold &\CheckmarkBold \\
laboratory 208 & \XSolidBold &\XSolidBold &\CheckmarkBold \\
any language & \XSolidBold &\XSolidBold &\CheckmarkBold \\
\hline
\end{tabular}
\begin{itemize}
    \item \textit{Expl.} stands for exploration, 
    \item \textit{Detect.} denotes detection modules\cite{sam,grounding_dino}. 
    \item \textit{Recent} refers to VLM-based works\cite{kuang2024openfmnav,vlfm}.
\end{itemize}

\end{table}

In this paper, for open goal navigation, we propose NavVLM, an open language navigation framework utilizing common sense reasoning ability of VLM for any language-defined goal. NavVLM is not an object-centric framework and thus has the capability of open goal navigation. Our framework leverages a small open-source VLM, capable of mobile device deployment, to achieve state-of-the-art performance. Our framework advances previous navigation systems towards open goal languages in various unknown indoor scenes. We demonstrate the effectiveness performance though both simulation and real-world validation, due to the sparsity of open goal navigation datasets, can only present cases for non-specific goals in Figure \ref{fig:episodes}. 
The difference of language goal capability is in Table \ref{tab1}.

\begin{figure*}[htbp]
    \centering
    
    \includegraphics[width=0.9\linewidth]{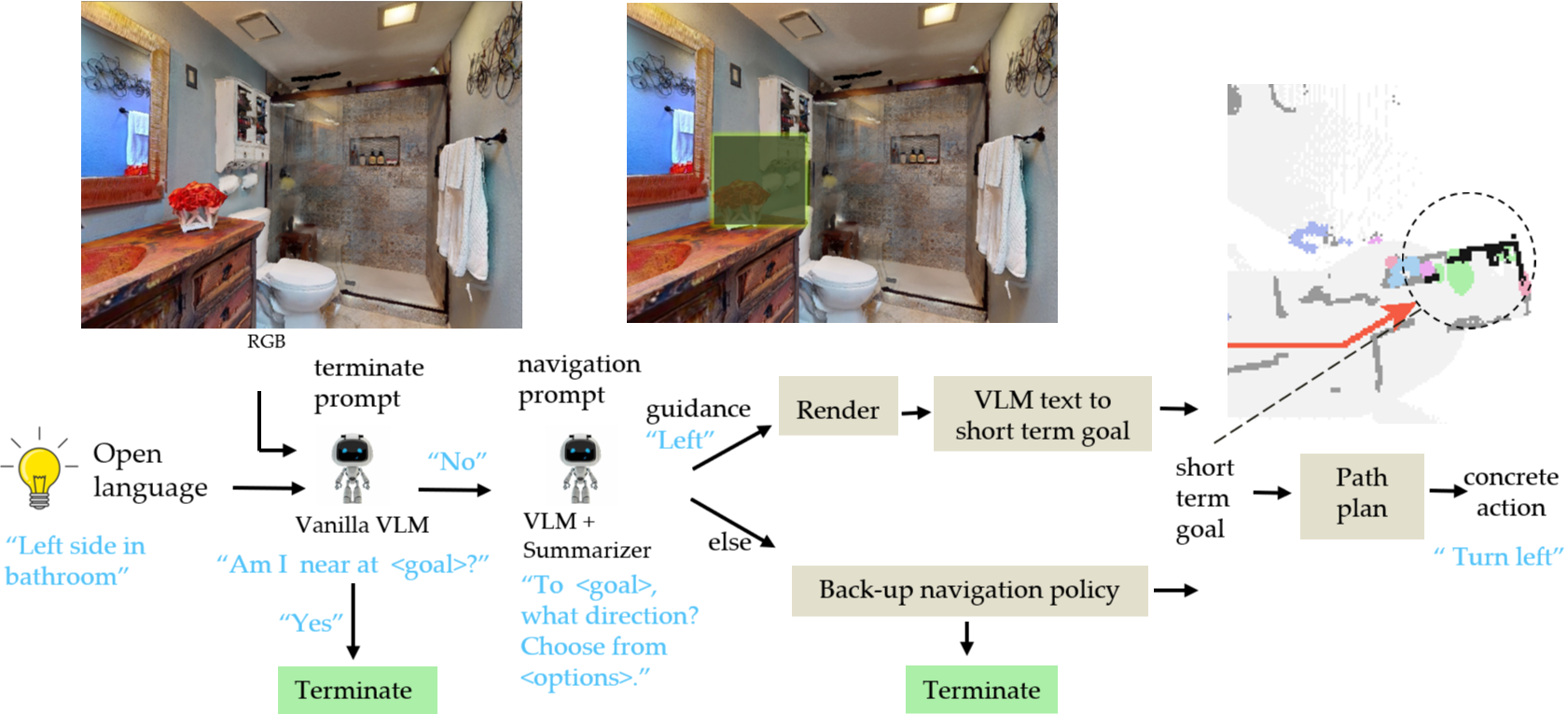}
    \caption{The overall framework. At each step, framework use textual prompts to provide navigation guidance based on the current observations. After projection of text guidance, the robot will go to the target area given by VLM. In this process, the VLM perceives information and acts as a high level intelligent commander, taking control of navigation.}
    \label{process}
\end{figure*}
\section{Related Works}
This section introduces several prominent recent intelligent object-oriented navigation works, as open goal navigation has not been explored in the previous works.

Detection-based works, which detects target objects along a classic exploration trajectory, are intuitive for ObjNav. \cite{chang2023goatthing} utilizes SuperClue\cite{superglue} and CLIP\cite{clip}, successfully extending the goal modality to images. However, the navigation goals are limited to specific targets, and the navigation policy is nearest frontier exploration\cite{frontier}, which struggles to exhibit satisfying intelligent exploration. \cite{conceptgraph} is an effective graph scene navigation system, but it is a two-stage framework that relies on pre-detection for later language goal query. Although the navigation is intelligent, the pre-exploration process is expensive when faced with unknown large environment. In addition, it is object-centric, and may fail on abstract places in open goal navigation settings.

Similarity-based approaches\cite{vlfm} utilize BLIP\cite{blip2}  to compute score between the RGB observation ahead and the language prompt to rank frontier exploration points. The language template of the goal must be specific such as ``there seems to be \textless{}goal\textgreater{} ahead", and the approach struggles to exhibit satisfying intelligence as it is fundamentally a similarity score-based method with less model expressivity. 

Pre-training navigation large VLMs\cite{navi_llm,embodied_generalist} require substantial inputs such as point clouds, point cloud features, and historical latent embeddings, which are resource intensive for edge devices. Recent learning based models\cite{navid,navid4d,opennav} can follow open spatial and temporal human instructions; however, they are focused on VLN-CE tasks where the instructions contain explicit environmental priors, which contradicts the principle of Open Goal Navigation. Diffusion-based models\cite{nomad,navidiffusor} can predict and generate local navigation waypoints on ego-centric observation but suffer from generalization problem to versatile navigation goals and environments.

Training-free VLM enhanced navigation works are also proposed. The object-centric method by \cite{kuang2024openfmnav} employs large models as a frontier scorer to rank frontiers based on the language goal, but it can not handle non-object goals such as ``middle of the hall". \cite{vlmnav} uses VLM to select next waypoint proposed by depth calculator based on vision through VQA to perform intelligent navigation to the goal, but this point-proposing-selecting scheme can easily cause collision as their ablation shows that simulation sliding setting has a huge impact on navigation success. With fixed positions in a closed scene, \cite{voro} uses VLMs to describe areas and build a room graph, and uses only language modalities through entire navigation, which sacrifices the open exploration capability that VLMs inherently possess. In \cite{shah2023lfg}, the VLM serves as a high-level planning system, but it requires multi-round dialog with the user. Similarly, in \cite{findthis}, to achieve good performance, the work also requires multi-round dialogs with a user to fix and adjust navigation. 

We propose NavVLM to further generalize ObjNav to open goal navigation, while our framework effectively addresses the issues encountered by previous models.
\begin{itemize}
    
    \item \textbf{Effective and intelligent navigation.} The VLM serves as a logical directional guider capturing various clues for effective navigation. For instance, for the goal ``cooker", the agent is more likely to head directly to the kitchen once it catches a distant glimpse of the kitchen. This is because NavVLM can logically infer, based on common consensus, that cookers are most likely to be found in kitchens.  Therefore, in both object-specific or open goal navigation tasks, NavVLM can achieve effective navigation.
    \item \textbf{Promising fully versatile language goal.} Our framework perform well in ObjNav, and promisingly handle open goal navigation tasks -- towards go to any language, where NavVLM framework can exploit VLMs capability to guide robot and navigate to the language open goal such as ``a corner", or abstract ``the middle of hall", as is not an object-centric framework.
    \item \textbf{Environment robust.}
    NavVLM enables autonomous open goal navigation without requiring human intervention or detailed instructions containing environmental priors, and operates independently of third-party camera overviews. It empowers the robot to navigate across large indoor areas without being restricted to a few rooms.
    
\end{itemize}



\section{NavVLM Framework}
\begin{figure}
\centering
\includegraphics[width=0.5\textwidth]{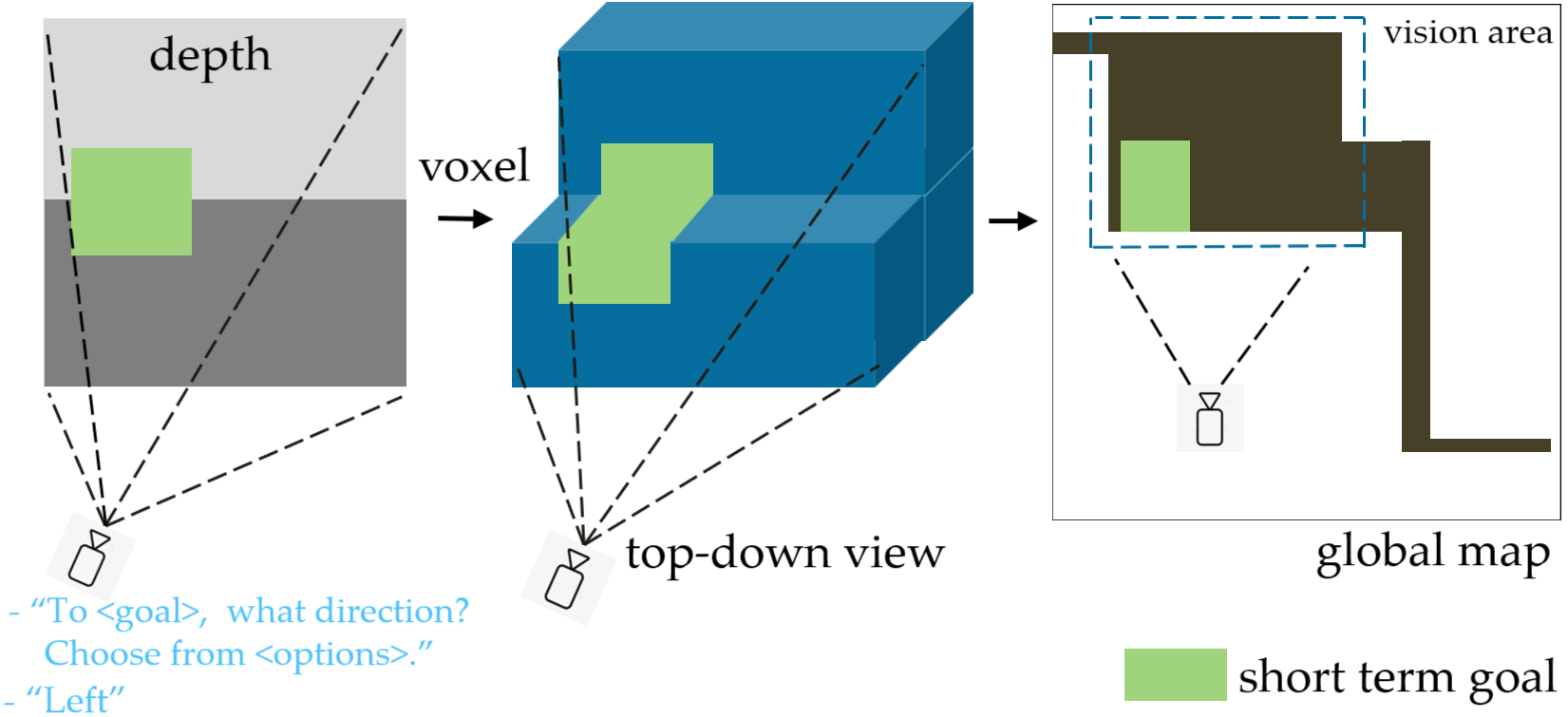}
\caption{VLM text to short-term goal} 
\label{text2act1-pic}
\end{figure}

\begin{figure}
\centering
\includegraphics[width=0.5\textwidth]{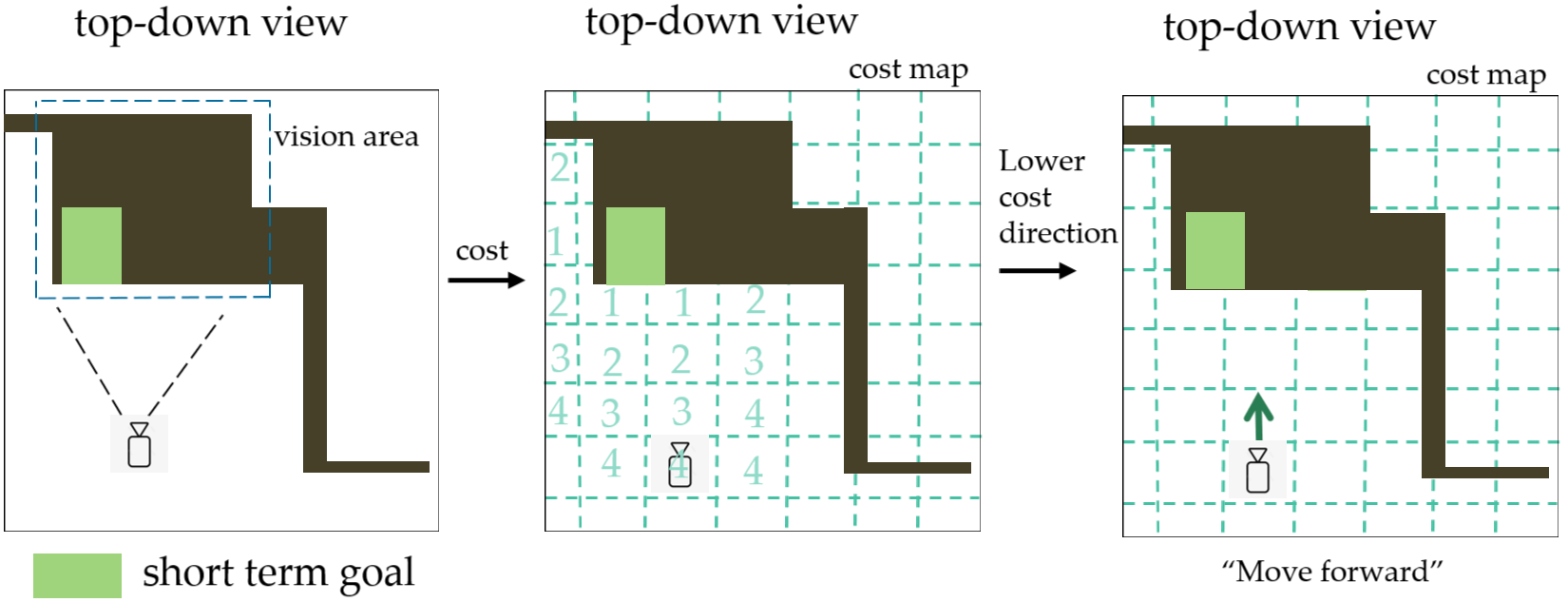}
\caption{Path plan overview} 
\label{path_plan}
\end{figure}

This section explains the components of our framework. The overall process is shown in Figure \ref{process} and a detailed demonstration of the process is provided in Algorithm \ref{alg}. As a huge control system, this process shows the key part of the process. \textit{VLM text to short term goal} module is explained in Figure \ref{text2act1-pic}, and \textit{path plan} module is explained in Figure \ref{path_plan}.

\subsection{Perceiving Information from Environment} \label{prompt}
The agent  receive an observation (RGB-D and pose) from the environment in each step. We utilize VLM to perceive these information with two different prompts. We firstly ask VLM whether the current observation satisfies termination in prompt such as ``Based on the current observation, is the \textless{}goal\textgreater{} in the front? ". 

Then together with current observation and top-down map with history trajectory, we ask VLM to provide directional information such as ``Based on the current observation, to find \textless{}goal\textgreater{}, which direction should you go, choose from \textless{}direction options\textgreater{}.". \label{termination} We refer the answer of VLM as \textbf{text guidance}, which is then converted into concrete action by action planner.

\subsection{VLM Text Guidance to Short Term Target Area} 
\label{text2act1}
The VLM text guidance is summarized as only directional options such as ``left, right, go straight" by LLM text summarizer (in our case the VLM itself with different prompt). These are just text directional guidance rather than concrete actions for the robot. To get concrete action, firstly, we render corresponding areas based on guidance in the observation image to convert it onto image area. 

For example, render the center-left area in the observation image when the VLM suggests going ``left", etc. Secondly, assign VLM guidance area in the image onto the voxel of current observation in corresponding location based on depth information. Thirdly, we used the top-down view of the voxel as a navigation map for path planning, and project the VLM guidance areas regarded as the ``short-term target area" onto the map.




\subsection{Path Planning towards Short Term Target Area} 
\label{text2act2}
Path planning involves moving the agent from current location to target area while avoiding obstacles. The short-term target area is provided by upper stream modules (mostly by VLM), we use fast marching method (FMM)\cite{ffm}-based path planning module to reach the short-term target area. Specifically, given the obstacle map built by depth, after FMM calculates the cost to the target areas, the robot steps to neighbouring area whose position hold a lower cost than the current and gives the concrete action such as ``turn left, turn right, move forward".

\begin{algorithm}[htbp]
\caption{Pseudocode of Overall Framework}
\label{alg}
\KwData{prompt for termination decision $p_{term}$,\\ prompt for VLM guidance $p_{nav}$,\\image with guidance $i\tilde{m}g$, top-down map $map$, short-term goal area $stg$, VLM text answer $ans$}

\For{$t \gets 1$ \textbf{to} $T$}{
    $(img, depth) \gets \text{GetObservation}()$;
    
    
    \If{$ 
    \text{VLM}(img, p_{term}) \lor \text{BackupTerm}(img, depth)$}{
    $done \gets \textbf{True}$;
    
        \textbf{break};
    }
    
    $ans \gets \text{VLM}(img, p_{nav})$;
    
    $guide \gets \text{Summarize}(ans)$;
    
    \eIf{ $guide \in 
    \{\text{left}, \text{right}, \text{forward}\}$}{
        
       $i\tilde{m}g \gets \text{Render}(img, guide)$;
       
       $voxel \gets \text{Voxelize}(i\tilde{m}g, depth)$;
       
       $map, stg \gets \text{TopDown}(voxel)$;
    }{
    
        $map, stg \gets \text{BackupNav}(img,depth)$;
        
    }
    
    $map \gets \text{Update}(d, map)$;
    
    $action \gets \text{PathPlan}(map, stg)$;
    
    $\text{StepAction}(action)$
}

\end{algorithm}
\subsection{Back-up Navigation Policy}
If the VLM does not provide valid information or explicitly commands ``explore more", the navigation process temporarily reverts to the back-up navigation strategy. In our case, we used nearest frontier exploration\cite{frontier} with detection. This does not play a key role as VLM mainly dominates the navigation proven by ablation study experiment at Figure \ref{fig:param_and_involvement}(b).

\subsection{Navigation Termination} 
Besides for max-step termination, we use the following termination.

As the VLM only provides short-term target area, reaching the VLM target area does not mean the end of navigation. So in addition, a determination prompt is used at every step to let VLM judge whether to stop the navigation based on current observation. 

Our backward compatible framework also accept termination from back-up navigation system, which uses detection to locate goal object and use its location as target area. If the robot reaches the target area, navigation is terminated.

\begin{table*}
\caption{Performance Comparison}
\label{tab:performance}
\centering

\begin{tabularx}{\textwidth}{XXX|cc|cc|cc}
\toprule

\multicolumn{1}{l}{Approach}&\multicolumn{1}{l}{Training-free}&\multicolumn{1}{l}{Open goal}&\multicolumn{2}{c}{Gibson} & \multicolumn{2}{c}{HM3D} & \multicolumn{2}{c}{MP3D} \\
\cmidrule(lr){4-5} \cmidrule(lr){6-7} \cmidrule(l){8-9}
 &  & & SPL↑ & SR↑  & SPL↑ & SR↑ & SPL↑ & SR↑ \\
\midrule
PONI\cite{poni}  & \XSolidBold & \XSolidBold & 41.0 & 73.6  & -&- & 12.1 & 31.8 \\
PIRLNav\cite{pirlnav} & \XSolidBold& \XSolidBold & - & - & 27.1 & \textbf{64.1} & - & - \\
RegQLearn\cite{gireesh2022objectgoalnavigationusing} & \XSolidBold & \XSolidBold & 31.3 & 63.7  &- &- & - &-\\
SemExp\cite{objnav} & \XSolidBold& \XSolidBold & 33.9 & 65.7 & - & - & - & - \\
ZSON \cite{majumdar2022zson} & \XSolidBold & \XSolidBold & - & - & 12.6 & 25.5 & 4.8 & 15.3 \\
CoW \cite{gadre2022cow}& \CheckmarkBold& \XSolidBold & - & - & - & - & 22.3 & 39.2 \\
SemUtil\cite{chen2023traindragontrainingfreeembodied} & \CheckmarkBold & \XSolidBold& 40.5 & 69.3 & - & - & - & -  \\
OpenFMNav\cite{kuang2024openfmnav} & \CheckmarkBold& \XSolidBold & -& - & 24.4 &54.9 & -& -\\
VLFM\cite{vlfm}  & \CheckmarkBold & \XSolidBold & 52.2 & \textbf{84.0}& 30.4 & 52.5 & 17.5 & 36.4\\
PIVOT\cite{pivot} & \XSolidBold & \XSolidBold &- &-  &10.6&24.6&- &- \\
VLMNav\cite{vlmnav}& \CheckmarkBold & \XSolidBold &- &-  &21.0& 50.4&- &-  \\
Ours  & \CheckmarkBold & \CheckmarkBold &  \textbf{56.4} & 72.3 & \textbf{33.5} & 48.0* & \textbf{27.9} & \textbf{40.0} \\
\bottomrule
\end{tabularx}
    \textit{*} with 30B VLM we can achieve 56.5\% SR.

\end{table*}

\begin{figure*}
    \centering
    \includegraphics[width=1\linewidth]{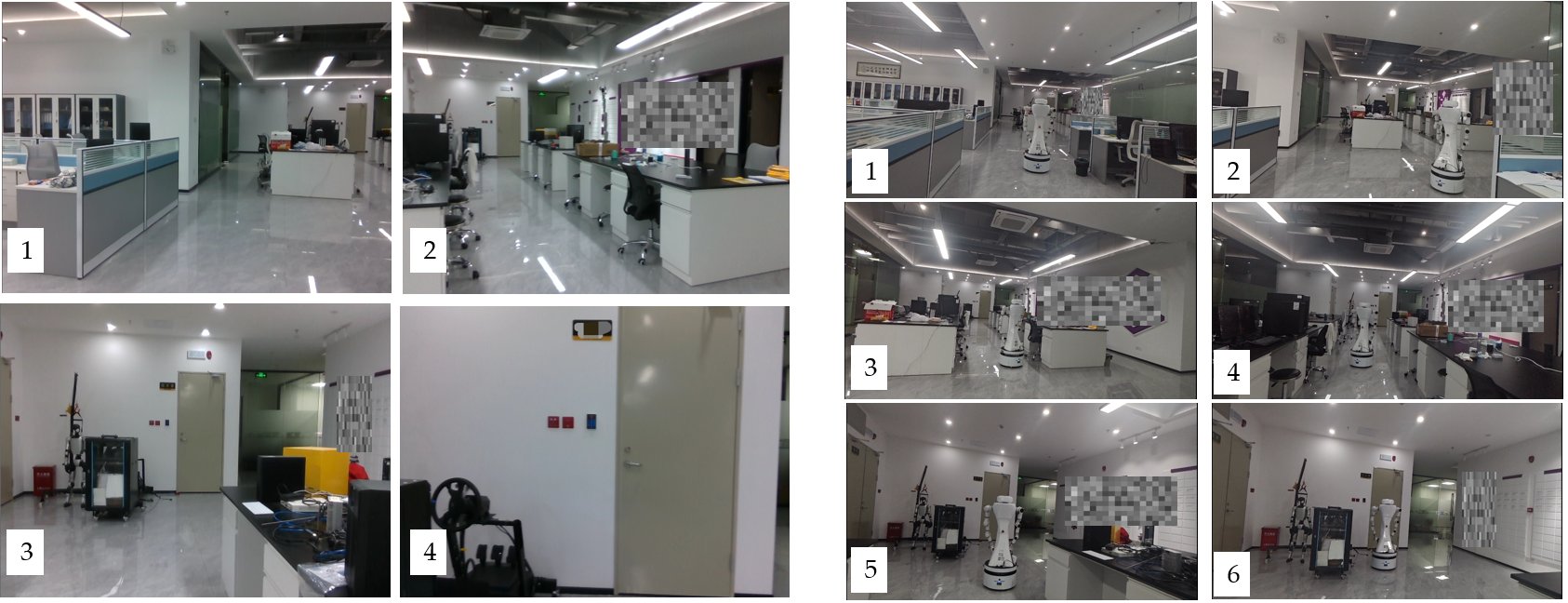}
    \caption{Real-world robot observation (left) and third-person view (right), ordered as numbered.}
    \label{fig:rw_obs}
\end{figure*}

\section{Simulation Experiment}
\label{sec:experiment}
\subsection{Experiment Setup}
Evaluation task is object-oriented navigation, we set the robot in unknown environments in different scenes in Gibson\cite{gibson}, HM3D\cite{hm3d}, and MP3D\cite{mp3d} to conduct object-oriented navigation. We selected minicpm-v2.6\cite{minicpm} as the cognitive VLM of the agent, with comparison on larger InterVL2-5\cite{internvl} series. We implemented the experiment via habitat\cite{habi3} on one GTX4080 server. The concrete actions included moving forward 0.25 meters, turning right 30 degrees, turning left 30 degrees, and terminating. We set max-step limit as 500. We used SPL and SR\cite{embodied_eval} as evaluation metrics. Note the upper bound value of SPL is SR when every exploration trajectory is optimal path. Performance comparisons are presented in Table \ref{tab:performance}. 

We adopt the object-specific navigation task for evaluation, as there currently lacks a navigation dataset specifically designed for non-object-centric navigation. Later in the paper, we also provide examples of open goal language navigation in Figure \ref{fig:episodes}. Furthermore, using this task facilitates comparison with existing baseline methods.

\subsection{Performance Analysis} 
SPL measures the average possibility of the agent finding the optimal path, which is the geodesic shortest path from the starting position to the goal position. SR measures the likelihood that the agent successfully navigates to the language goal. Our framework outperforms all baselines in SPL and achieves competitive SR scores across all datasets.

\textbf{Compared with reinforcement learning methods} (baselines before ZSON), we achieved the competitive performance in both SPL and SR across three datasets. We attribute this improvement to VLM sense ability over RL-based action policy on limited scenes and limited goal vocabularies, as the common sense reasoning ability VLM itself contains can capture clues in the observation and guide our navigation in more generalized scenes. 

\textbf{Compared with VLM-based methods} such as OpenFMNav, PIVOT, and VLMNav, we achieved the highest SPL on HM3D, demonstrating the most effective navigation performance among them. By introducing simple and flexible modules to utilize VLM abilities, we can better transfer the commonsense reasoning capabilities of VLMs to navigation effectiveness.

\textbf{Compared with other baselines} including SemUtils, VLFM, CoW, our framework remains state-of-the-art in SPL across all datasets, as VLM can utilize common sense comprehensively that capture environment hints to guide robot to the goal more effectively.

\textbf{Our framework perform effectively} with only 500 steps as maximum step, while baselines such as VLFM\cite{vlfm} could take up to 50000 steps. Smaller max-step may compromise our SR but it also showcases competitive navigation capabilities across a variety of dataset scenes, highlighting its robustness, generalization, and adaptability.

\subsection{Open Goal Navigation Episodes}
\begin{figure}[t]
\centering

\begin{tabular}{c}
\includegraphics[width=1\linewidth]{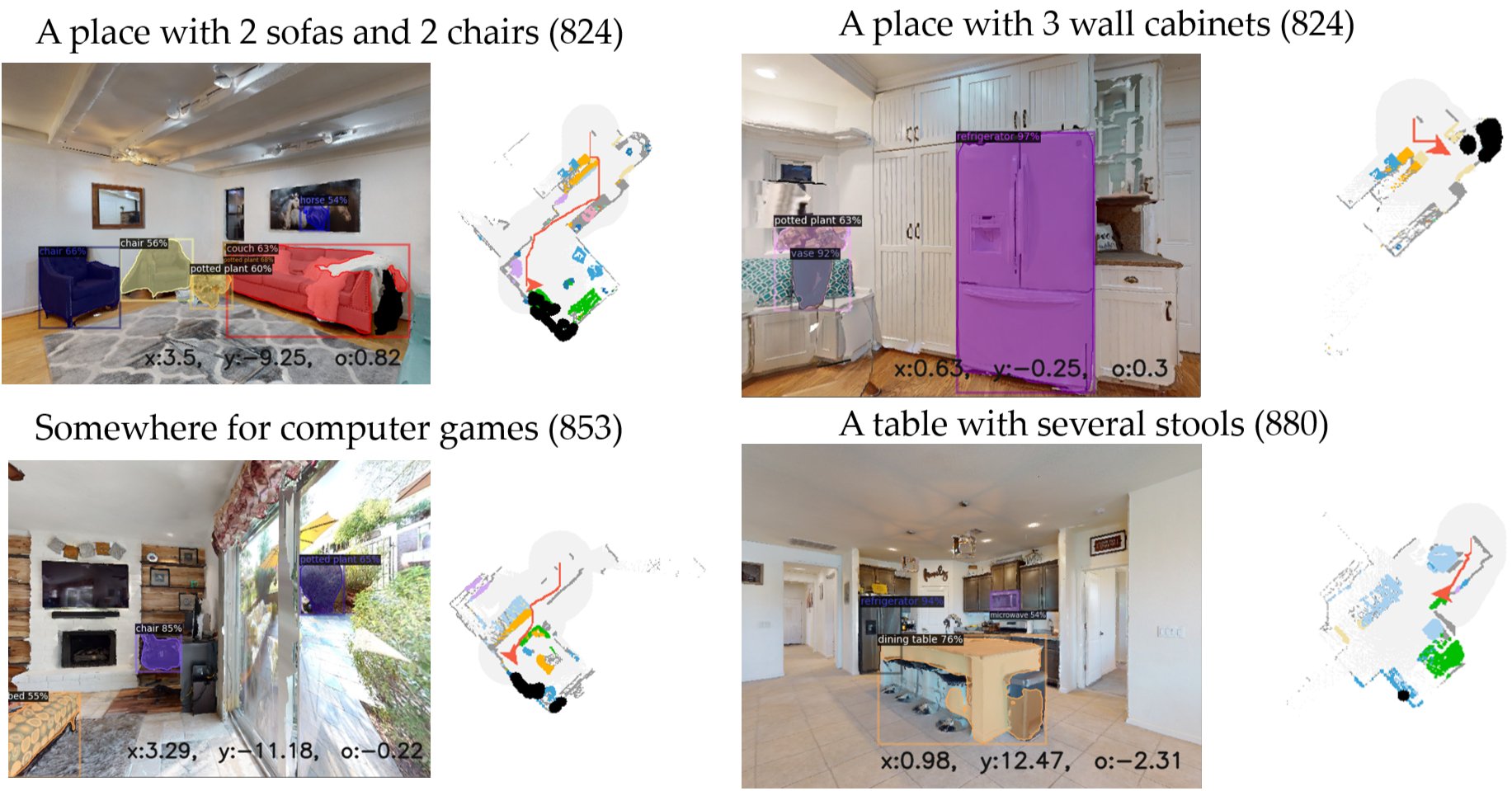} \\
(a) \\[1em]
\includegraphics[width=1\linewidth]{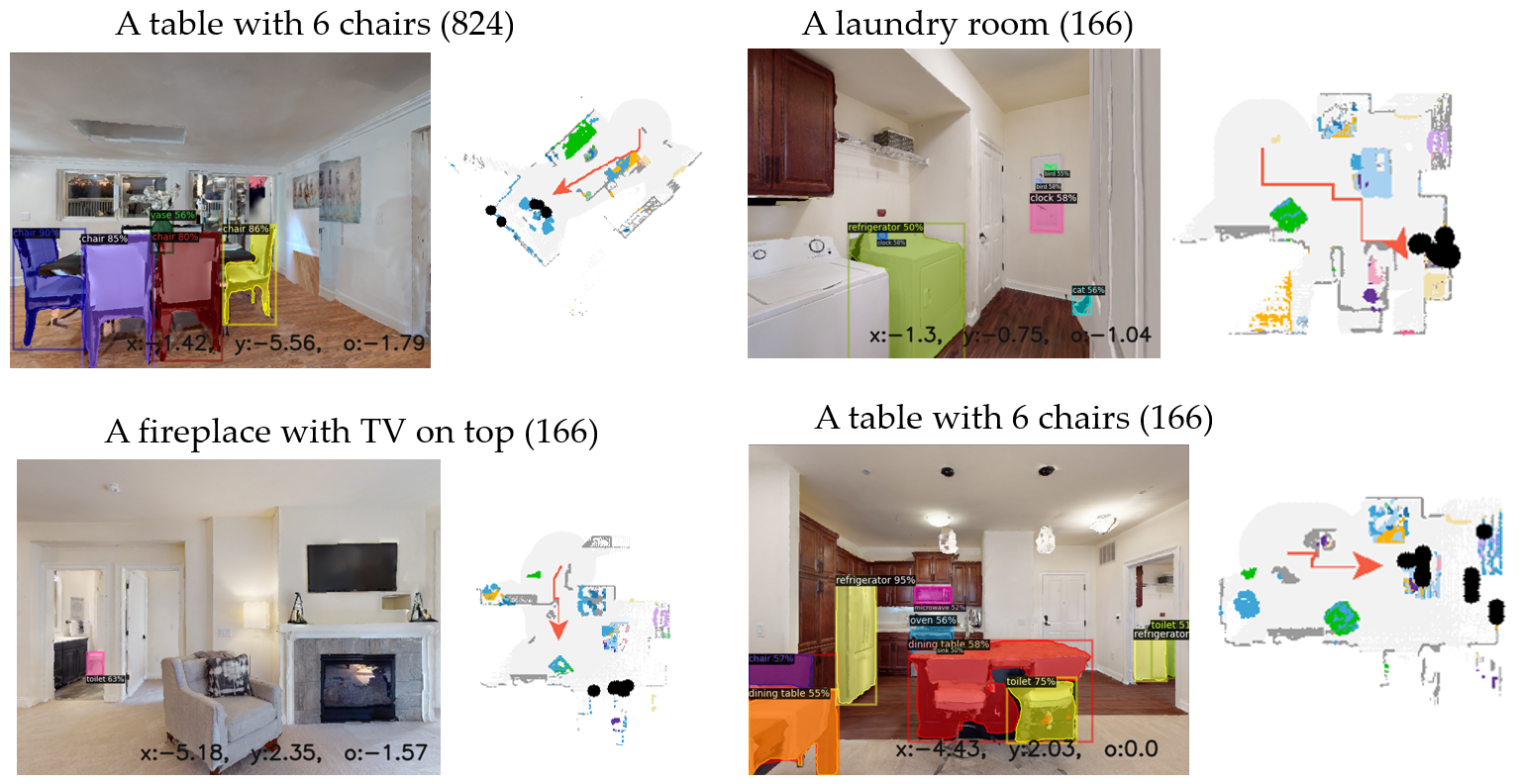} \\
(b)
\end{tabular}
\caption{Open goal navigation with scene id. }
\label{fig:episodes}
\end{figure}

NavVLM has the ability of open goal navigation. Comparison with object-centric navigation experiments demonstrates the effectiveness of our framework and this section showcases our framework's open goal navigation capability. However, due to the lack of human language navigation dataset and the challenges generating such data, we could only present a sufficient number of episode cases in Figure \ref{fig:episodes} to show our framework's open goal navigation capability. 

These episodes feature relatively short trajectories, making them easier to visualize. Since VLM termination can occur at any step when the VLM determines that the observation matches the language goal, the VLM guidance area (the black area in the figure) is not guaranteed to be within the actual goal area nor to be reached. 

Please note that the VLM guidance is generated from NavVLM framework and does not depend on detection bounding box. It is also implied from the fact the VLM guidance (black area) does not align with any object color. These detections help object-centric navigation but of no use in human language navigation. We intend to keep the bounding box in visualization to demonstrate NavVLM can cooperate with back-up system.

\begin{figure}
\centering
\includegraphics[width=0.4\textwidth]{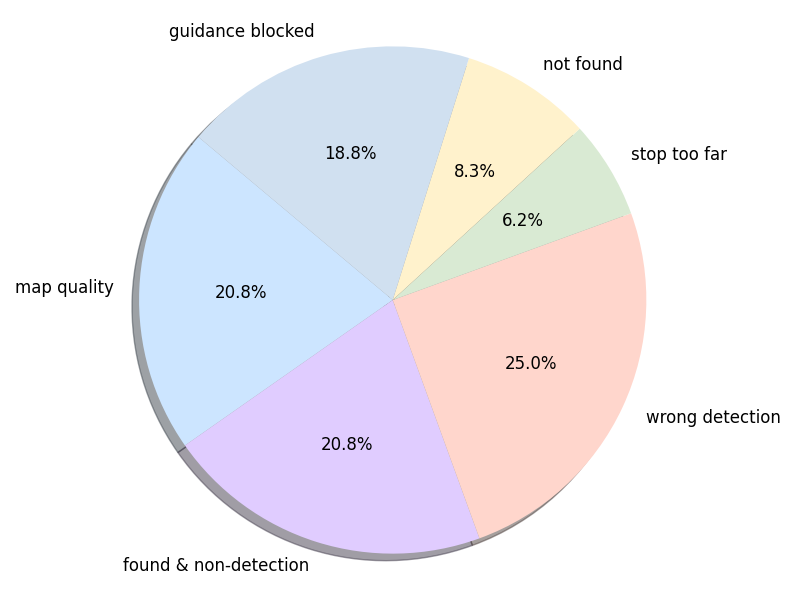}
\caption{Failure statistics in simulation} 
\label{failure}
\end{figure}

\subsection{Ablation Study}

\begin{figure}[t]
\centering

\begin{tabular}{c}
\includegraphics[width=1\linewidth]{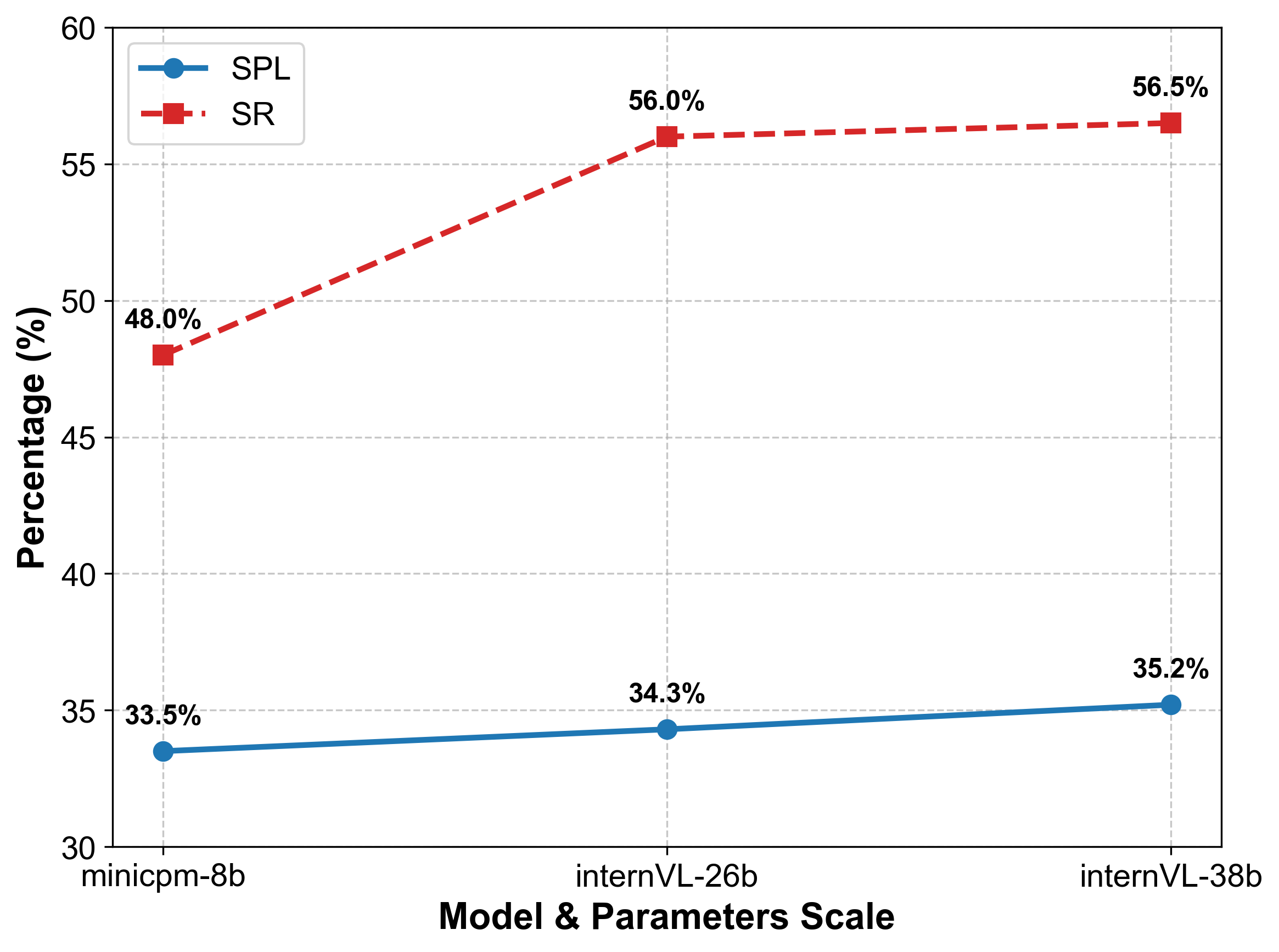} \\
(a) \\[1em]
\includegraphics[width=1\linewidth]{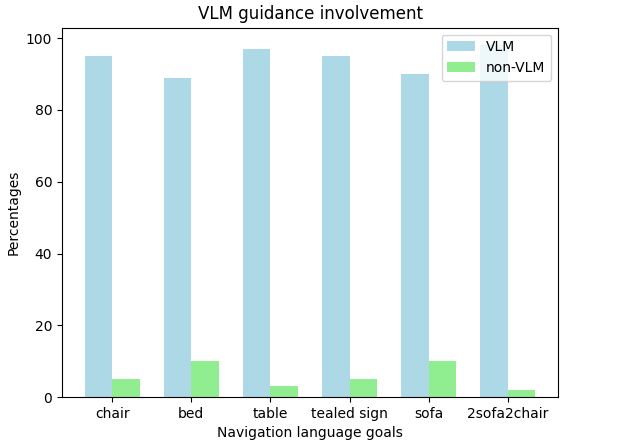} \\
(b)
\end{tabular}

\caption{(a) parameter scale ablation on HM3D. (b) VLM guidance dominates navigation.}
\label{fig:param_and_involvement}
\end{figure}

\textbf{As the VLM parameter scales up, how does the performance vary?} We compared SR, SPL performance in HM3D for frameworks in different scales of VLMs, in Figure \ref{fig:param_and_involvement}(a). Using larger models yields only marginal improvements. The incremental gains do not justify the largely increased computational demands compared to employing smaller VLMs.

\textbf{What factors contribute to the failures?}
Figure \ref{failure} demonstrates failure statistics.
The main cause of failure contains four parts. \textit{Wrong detection} come from back-up navigation detection caused by misunderstanding of the observation in particular view points. \textit{fonud but non-stop} refers to detection or VLM fail to recognize items and pass by. \textit{map quality} refers to the incompletion of the 3D scene, such as geometry distortion and error spatial perspective caused by wall absence. Similar to \cite{vlmnav}, VLM ego-centric selection schemes rely on the good completion of environment scene.  \textit{guidance blocked} is inaccurate projection from VLM guidance to map, an inherent limitation of NavVLM. \textit{stop too far} is from wrong VLM modules termination or inappropriate path planning. \textit{not found} is from max-step termination.


\textbf{How much does VLM guidance involves the navigation?} We document the ratio of concreate actions taken based on VLM guidance, which accounts for up to 90\% of 200 navigation process across episodes in HM3D. In contrast, back-up navigation methods contribute to only 10\% of the process. This highlights that the significant improvements in SR and SPL are primarily attributed to the advancements in VLM. Figure \ref{fig:param_and_involvement}(b) selects several navigation cases from both object-centric goal and any language goal in HM3D 824 scene to demonstrate that VLM guidance truely dominates the navigation process. As a matter of axis space we use \textit{tealed sign} short for \textit{tealed colored DREAM sign} and \textit{2sofa2chair} for \textit{some place that has 2 sofas and 2 chairs}.

\begin{table}[t]
\centering
\caption{Real-world ablation on goal}
\label{tab:rw_ablation}

\begin{tabular}{|c|c|c|}
\hline
Open Lang.    & w.o. VLM & w. VLM(Ours)\\ \hline
Succ/All & 0/5 & 3/5 \\ \hline
SR       & 0\% & 60\% \\ \hline
TV    & w.o. VLM & w. VLM(Ours) \\ \hline
Succ/All & 3/5 & 4/5 \\ \hline
SR       & 60\% & 80\% \\ \hline
SPL       & 36.4\% & 54.1\% \\ \hline
\end{tabular}
\begin{itemize}
    \item \textit{door with yellow black chip room sign aside} (Open Lang.) and \textit{TV}    

\end{itemize}

\end{table}

\begin{figure}
    \centering
    \includegraphics[width=0.8\linewidth]{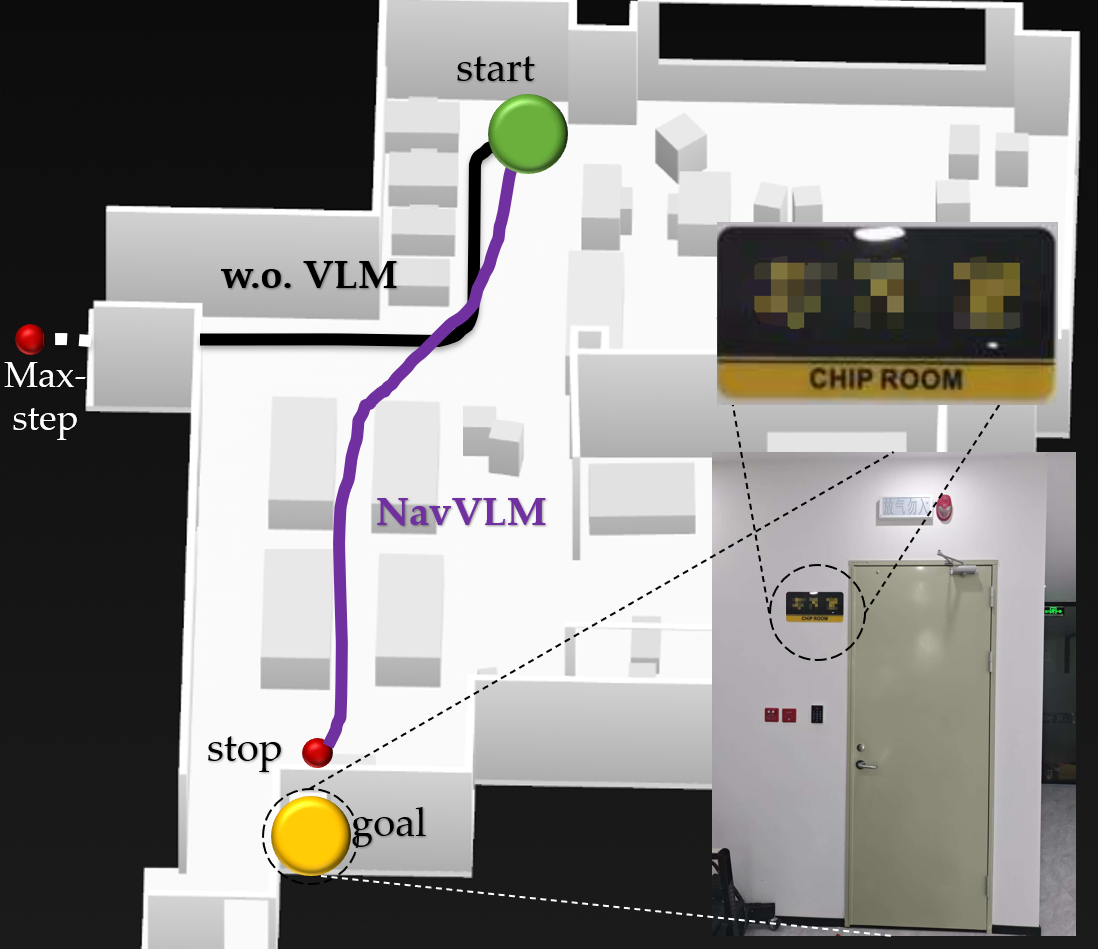}
    \caption{Real-world validation on goal \textit{door with yellow black chip room sign aside} (Open Lang.). In open goal navigation, NavVLM can navigate successfully and terminate at proper position, while the \textit{w.o. VLM} can not reach target language goal.}
    \label{fig:ablation}
\end{figure}

\section{Real-world Robot Validation}
We used a human-like half-body RealMan robot with a wheeled base for navigation tasks.An Intel RealSense RGB-D sensor was mounted on the robot on Ubuntu 20.04 system, from which we directly obtained point cloud data using the sensor’s built-in post-processing. We defined three atomic actions: ``move forward 0.5 meters", ``turn left 30 degrees", and ``turn right 30 degrees". The robot's pose changes were calculated using IMU odometry. We manually measured trajectory length. The real-world max-step is 200.

The experiments were conducted in our lab office. We tested ``door with yellow black chip room sign aside" as human language navigation goal, which is out of detection domain and ``TV" as in-domain navigation goal. We conducted five episodes, and the results are summarized in the following Table \ref{tab:rw_ablation}, and the observations are in Figure \ref{fig:rw_obs}.


We conducted validation ablation experiments, navigating to target language without VLM guidance(\textit{w.o. VLM} in Table \ref{tab:rw_ablation}). As ``door with yellow black chip room sign aside" is not in the detector domain, thus it fails at all attempts, while navigation to TV is not effected. However, the SPL of our framework is better than \textit{w.o. VLM}.


\section{Conclusion}
In this paper, we propose NavVLM, a training-free framework of effective ObjNav and even for open goal navigation in unknown environments, enabling intelligent navigation towards any language goals.
The open-source VLM of agent plays as a cognitive core and constantly give exploration guidance throughout the navigation towards specific goal and non-specific language goal. NavVLM achieves autonomous intelligent navigation without dependencies on human instructions or environmental prior. Given the lack of non-object navigation data, we report the state-of-the-art performance on SPL and competitive SR in ObjNav tasks in multiple vivid simulation scenes, and present a large number of successful open goal navigation episodes, demonstrating NavVLM's open-set language navigation capability. We validated our framework and conducted ablation studies on a real robot in indoor environments, demonstrating its practical effectiveness in real-world settings.





\section*{multimedia}
We provide real world robot episode in multimedia.
~\vspace{8cm} 


\bibliographystyle{IEEEtranS}
\bibliography{IEEEexample}

\end{document}